\documentclass[a4paper,10pt,conference]{IEEEtran}

\usepackage{amsmath,amsthm,amsfonts,amssymb,float,marginnote}
\usepackage{todonotes}
\usepackage{subcaption}
\usepackage{ dsfont }
\usepackage{graphicx}
\usepackage{epsfig}
\usepackage{color, lipsum}
\usepackage{tikz, tikz-qtree}
\usepackage{multicol, multirow}
\usepackage{algorithm}
\usepackage[noend]{algpseudocode}

\newtheorem{theorem}{Theorem}

\newtheorem{definition}{Definition}

\newtheorem{remark}{Remark}

\newcommand{\AM} [1]{}

\DeclareGraphicsExtensions{.png}
\DeclareGraphicsExtensions{.ps}

\def\Gr{\mathcal{G}}

\def\R{{\mathbb{R}}}

\def\N{{\mathbb{N}}}

\def\Bernoulli{\mathrm{Bernoulli}}
\def\epsres{\epsilon_{res}}

\def\SBM{\mathrm{SBM}}
\def\Param{\mathcal{P}}

\def\Act{\mathcal{A}}

\algnewcommand{\LeftComment}[1]{\Statex \(\triangleright\) #1}

\DeclareGraphicsExtensions{.pdf}

\begin{document}

\title{The Power of Graph Convolutional Networks to Distinguish Random Graph Models: Short Version}

\author{
\IEEEauthorblockN{Abram Magner}
\IEEEauthorblockA{ 
        University at Albany, SUNY \\ 
        Albany, NY, USA\\
        Email: amagner@albany.edu
}
\and 
\IEEEauthorblockN{Mayank Baranwal}
\IEEEauthorblockA{
        University of Michigan \\
        Ann Arbor, MI, USA \\
        Email: mayankb@umich.edu
}
\and
\IEEEauthorblockN{Alfred O. Hero III}
\IEEEauthorblockA{
        University of Michigan \\
        Ann Arbor, MI, USA \\
        Email: hero@eecs.umich.edu
}
}

%
\date{}
\maketitle

\begin{abstract}
    Graph convolutional networks (GCNs) are a widely used method for
    graph representation learning.  
    We investigate the power of GCNs, as a function of their number of
    layers, to distinguish between different
    random graph models on the basis of the embeddings of their sample
    graphs.  In particular, the graph models that we consider arise from
    graphons, which are the most general possible parameterizations of
    infinite exchangeable graph models and which are the central objects of study
    in the theory of dense graph limits.  We exhibit an infinite class
    of graphons that are well-separated in terms of cut distance and are indistinguishable by a GCN with nonlinear activation functions coming from
    a certain broad class if its depth is at
    least logarithmic in the size of the sample graph.
    These results theoretically match empirical observations of several
    prior works.  
    Finally, we show a converse result that for pairs of graphons satisfying a degree profile separation property, a very simple GCN architecture suffices for distinguishability.  To prove our results, we exploit a connection to random
    walks on graphs.

\end{abstract}

\section{Introduction}
\label{Introduction}

In applications ranging from drug discovery \cite{sun2019graph} and design to proteomics \cite{randic2002comparative} to neuroscience \cite{sporns2003graph} to social network analysis \cite{barnes1983graph}, inputs to machine learning methods
take the form of graphs.  In order to leverage the empirical success of deep
learning and other methods that work on vectors in finite-dimensional Euclidean spaces for supervised learning tasks in this domain, a plethora of graph representation learning schemes have
been proposed and used~\cite{Hamilton2017RepresentationLO}.
One particularly effective such method is the \emph{graph convolutional network}
(GCN) architecture~\cite{kipfwelling,vandergheynst}.  
A graph convolutional network works by associating with each node of an input
graph a vector of features and passing these node features through a sequence of
\emph{layers}, resulting in a final set of node vectors, called node embeddings.
To generate a vector representing the entire graph, these final embeddings are 
sometimes averaged.  Each layer of the network consists of a graph diffusion step,
where a node's feature vector is averaged with those of its neighbors; a feature
transformation step, where each node's vector is transformed by a weight matrix;
and, finally, application of an elementwise nonlinearity such as the ReLU or
sigmoid function.  The weight matrices are trained from data, so that the metric 
structure of the resulting
embeddings are (one hopes) tailored to a particular classification task.

While GCNs and other graph representation learning methods have been successful
in practice, numerous theoretical questions about their capabilities and the
roles of their hyperparameters remain unexplored.  In this paper, we give results on the
ability of GCNs to distinguish between samples from different random graph models.
We focus on the roles that the number of layers and the presence or absence
of nonlinearity play.
The random graph models that we consider are those that
are parameterized by \emph{graphons}~\cite{lovaszbook}, which are functions from
the unit square to the interval $[0, 1]$ that essentially encode edge density
among a continuum of vertices.  Graphons are the central objects of study in the
theory of dense graph limits and, by the Aldous-Hoover theorem~\cite{aldous-exchangeable} exactly parameterize the class of infinite
exchangeable 
random graph models -- those models whose samples are invariant in distribution 
under permutation of vertices.

\subsection{Prior Work}
A survey of modern graph representation learning methods is provided in~\cite{Hamilton2017RepresentationLO}.
Graph convolutional networks were first introduced in~\cite{vandergheynst}, and since then, many variants have been proposed.
For instance, the polynomial convolutional filters in the original work
were replaced by linear convolutions~\cite{kipfwelling}. Authors in \cite{ruiz2019gated} modified the original architecture to include gated recurrent units for working with dynamical graphs. These and other variants have been used in various applications, e.g.,~\cite{jepsen2019graph,coley2019graph,yao2019experimental,duvenaud2015convolutional}.

Theoretical work on GCNs has been from a variety of perspectives.
In~\cite{verma2019stability}, the authors investigated the generalization and stability properties of GCNs.  Several works,
including~\cite{morris2019weisfeiler,chen2019equivalence,xu2018powerful}, have drawn connections between the representation
capabilities of GCNs and the distinguishing ability of the \emph{Weisfeiler-Lehman} (WL) algorithm for graph isomorphism testing~\cite{Weisfeiler1968ReductionOA}.  These
papers drawing comparisons to the WL algorithm implicitly study the injectivity
properties of the mapping from graphs to vectors induced by GCNs.  However, they
do not address the metric/analytic properties, which are important in consideration
of their performance as representation learning methods~\cite{arorarepresentationlearning}.  Finally, at least one work has considered the performance of untrained GCNs on community detection~\cite{Kawamoto2018}. { The authors of that paper provide a heuristic calculation based on the mean-field approximation from statistical physics and demonstrate through numerical experiments the ability of untrained GCNs to detect the presence of clusters and to recover the ground truth community assignments of vertices in the stochastic block model. They empirically show that the regime of graph model parameters in which an untrained GCN is successful at this task agrees well with the analytically derived detection threshold. The authors also conjecture that training GCNs does not significantly affect their community detection performance. }

The theory of graphons as limits of dense graph sequences was
initiated in~\cite{LovaszGraphLimits} and developed by various
authors~\cite{BorgsChayes2008,BorgsChayesII}.
For a comprehensive treatment, see~\cite{lovaszbook}.

Several authors have investigated the problem of estimation of
graphons from samples~\cite{ChanAiroldi2014,gao2015,Klopp2019}.  Our work is complementary
to these, as our goal is to investigate the performance of a
\emph{particular} method on the problem of distinguishing graphons.

\subsection{Our Contributions}
We first establish a convergence result for GCN embedding vectors, which will give 
a lower bound on the probability of error of \emph{any} test that
attempts to distinguish between two graphons based on slightly perturbed $K$-layer GCN embedding 
matrices of sample graphs of size $n$, provided that $K = \Omega(\log n)$.  In particular, we exhibit a family of pairs of graphons that are hard for any test to distinguish on the basis of these embeddings.  This is the content of Theorems~\ref{thm:convergence-result} and \ref{thm:prob-err-lower-bound}.

We then show a converse achievability result in Theorem~\ref{thm:achievability} that says, roughly, that provided that the number of layers is
sufficiently large ($K = \Omega(\log n)$), there exists a \emph{linear} GCN architecture 
with a very simple sequence of weight
matrices and a choice of initial embedding matrix such that pairs of graphons whose expected 
degree statistics
differ by a sufficiently large amount are distinguishable from the noise-perturbed GCN embeddings of their sample graphs.  In other words, this indicates that the family of difficult-to-distinguish graphons alluded to above is essentially the \emph{only} sort of
case in which a nonlinear GCN architecture could be necessary (though, as Theorem~\ref{thm:prob-err-lower-bound} shows, for several choices of activation functions,
these graphons are still indistinguishable).

Our proofs rely on concentration of measure results and techniques from the theory
of Markov chain mixing times and spectral graph theory~\cite{LevinPeresWilmer2006}.

\subsubsection{Relations between probability of error lower and upper bounds}
Our probability of error lower bounds give theoretical backing to a phenomenon that
has been observed empirically in graph classification problems: adding arbitrarily
many layers (more than $\Theta(\log n)$) to a GCN can substantially degrade 
classification performance.  This is an implication of Theorem~\ref{thm:prob-err-lower-bound}. 
On the other hand, Theorem~\ref{thm:achievability} shows that this is \emph{not} always
the case, and that for \emph{many} pairs of graphons, adding more layers improves
classification performance.  We suspect that the set of pairs of graphons for which
adding arbitrarily many layers does not help forms a set of measure $0$, though this
does not imply that such examples never arise in practice.

The factor that determines whether or not adding layers will improve or degrade performance of a GCN in distinguishing between two graphons $W_0$ and $W_1$ is the distance between the stationary distributions
of the random walks on the sample graphs from $W_0$ and $W_1$. 
This, in turn,
is determined by the normalized degree profiles of the sample graphs.

An extended version of this paper is available on ArXiv~\cite{jsait2019}.

\section{Notation and Model}
\subsection{Graph Convolutional Networks}
We start by defining the model and relevant notation.  A $K$-layer
graph convolutional network (GCN) is a function mapping graphs to vectors over
$\R$.  It is
parameterized by a sequence of $K$
\emph{weight matrices} $W^{(j)} \in \R^{d\times d}$, $j \in \{0, ..., K-1 \}$, where $d \in \N$ is the \emph{embedding dimension}, a hyperparameter.
From an input graph $G$ with adjacency matrix $A$ and random walk matrix $\hat{A}$ (i.e., $\hat{A}$ is $A$ with every row normalized by the sum of its
entries),
and starting with an initial embedding matrix $\hat{M}^{(0)}$, the $\ell$th
embedding matrix is defined as follows:
\begin{align}
    \hat{M}^{(\ell)}
    =  \sigma(\hat{A} \cdot \hat{M}^{(\ell-1)} \cdot W^{(\ell-1)}),
    \label{GCNRecurrence}
\end{align}
where $\sigma:\R\to\R$ is a fixed nonlinear \emph{activation function}
and is applied element-wise to an input matrix.  An \emph{embedding vector} $\hat{H}^{(\ell)} \in \R^{1\times d}$ is then produced by averaging the rows of $\hat{M}^{(\ell)}$:
\begin{align}
    \hat{H}^{(\ell)}
    = \frac{1}{n} \cdot \mathbf{1}^T \hat{M}^{(\ell)}.
\end{align}

Typical examples of activation functions in neural
network and GCN contexts include the ReLU, sigmoid, and
hyperbolic tangent functions.
Empirical work has given evidence that the performance of GCNs on certain
classification tasks is unaffected by replacing nonlinear activation functions
by the identity~\cite{Wu2019SimplifyingGC}.  Our results lend
theoretical credence to this.

Frequently, $\hat{A}$ is replaced by either the normalized adjacency matrix
$D^{-1/2}AD^{-1/2}$, where $D$ is a diagonal matrix with the degrees of the
vertices of the graph on the diagonal, or some variant of the Laplacian matrix 
$D - A$.  For simplicity, we will consider in this paper only the choice of 
$\hat{A}$.

The defining equation (\ref{GCNRecurrence}) has the following interpretation:
multiplication on the left by $\hat{A}$ has the effect of replacing each node's
embedding vector with the average of those of its neighbors.  Multiplication on
the right by the weight matrix $W^{(\ell-1)}$ has the effect of replacing each
coordinate (corresponding to a feature) of each given node embedding vector with a 
linear combination of values of the node's features in the previous layer.

\subsection{Graphons}
In order to probe the ability of GCNs to distinguish random graph models
from samples, we consider the task of distinguishing random graph models
induced by graphons.  A graphon $W$ is a symmetric, Lebesgue-measurable
function from $[0, 1]^2 \to [0, 1]$.  To each graphon is associated
a natural exchangeable random graph model as follows: to generate a graph
on $n$ vertices, one chooses $n$ points $x_1, ..., x_n$ uniformly
at random from $[0, 1]$.  An edge between vertices $i, j$ is independent
of all other edge events and is present with probability $W(x_i, x_j)$.
We use the notation $G \sim W$ to denote that $G$ is a random sample
graph from the model induced by $W$.  The number of vertices will be
clear from context.

One commonly studied class of models that may be defined
equivalently in terms of sampling from graphons is the class of 
stochastic block models.  A stochastic
block model on $n$ vertices with two blocks is parameterized by
four quantities: $k_1, p_1, p_2, q$.  The two blocks of vertices
have sizes $k_1 n$ and $k_2 n = (1-k_1)n$, respectively.  Edges
between two vertices $v, w$ in block $i$, $i \in \{1, 2\}$, 
appear with probability $p_i$, independently of all other edges.
Edges between vertices $v$ in block $1$ and $w$ in block $2$
appear independently with probability $q$.  We will 
write this model as $\SBM(p_1, p_2, q)$, suppressing $k_1$.

An important metric on graphons is the \emph{cut distance}~\cite{jansongraphons}.  It is induced
by the cut norm, which is defined as follows: fix a graphon $W$.  Then
\begin{align}
    \| W \|_{cut} = \sup_{S,T} \left|\int_{S\times T} W(x, y) ~d\mu(x)~d\mu(y) \right|,
\end{align}
where the supremum is taken over all measurable subsets of $[0,1]$, and the integral is taken with respect to the Lebesgue
measure.  For finite graphs, this translates to taking the pair of subsets $S, T$ of vertices that has the maximum between-subset edge density.
The cut \emph{distance} $d_{cut}(W_0, W_1)$ between graphons $W_0, W_1$ is then defined as
\begin{align}
    d_{cut}(W_0, W_1)
    = \inf_{\phi} \|W_0 - W_1(\phi(\cdot), \phi(\cdot))\|_{cut},
\end{align}
where the infimum is taken over all measure-preserving bijections of $[0, 1]$.  In the case of finite graphs,
this intuitively translates to ignoring vertex labelings.
The cut distance generates the same topology on the space of graphons
as convergence of subgraph homomorphism densities (i.e., \emph{left convergence}), and so it is an important part of the theory of graph limits.

\subsection{Main Hypothesis Testing Problem}
We may now state the hypothesis testing problem under consideration.
Fix two graphons $W_0, W_1$.  A coin $B \sim \Bernoulli(1/2)$ is flipped,
and then a graph $G \sim W_B$ on $n$ vertices is sampled.  Next, $G$
is passed through $K=K(n)$ layers of a GCN, resulting in a matrix $\hat{M}^{(K)} \in \R^{n\times d}$
whose rows are node embedding vectors.  The graph embedding
vector $\hat{H}^{(K)}$ is then defined to be $\frac{1}{n}\mathbf{1}^T \hat{M}^{(K)}$.
As a final step, the embedding
vector is perturbed in each entry by adding an independent, uniformly
random number in the interval $[-\epsres, \epsres]$, for a parameter
$\epsres > 0$ that may depend on $n$, which we will typically consider to be $\Theta(1/n)$.  This results in a vector
$H^{(K)}$.  We note that this perturbation step has precedent 
in the context of studies on the performance of neural networks in the presence of numerical 
imprecision~\cite{numericalprecisionShanbhag}.  For our purposes, it will allow
us to translate convergence results to information theoretic lower bounds.

Our goal is to study the effect of the number of layers $K$ and presence or absence of nonlinearities on the 
representation properties of GCNs and probability of error of optimal tests $\Psi(H^{(K)})$ that are meant to 
estimate $B$.  Throughout, we will consider the case where $d=n$.  We will
frequently use two particular norms: the $\ell_{\infty}$ norm for vectors and
matrices, which is the maximum absolute entry; and the operator norm induced
by $\ell_{\infty}$ for matrices: for a matrix $M$, 
\begin{align}
    \| M \|_{op,\infty}
    = \sup_{v ~:~ \|v\|_{\infty}=1} \| Mv\|_{\infty}.
\end{align}
\AM{FINISH THIS!  Make sure that we're consistent in our norm notation!}

\section{Main Results}

To state our results, we need a few definitions.
For a graphon $W$, we define the
degree function $d_W:[0, 1]\to \R$ to be
\begin{align}
    d_W(x)  = \int_{0}^1 W(x, y)~dy,
\end{align}
and define the total degree function
\begin{align}
    D(W) = \int_{0}^1 \int_{0}^1 W(x, y) ~dx~dy.
\end{align}
We will assume in what follows that all graphons $W$ have the property that
there is some $\ell > 0$ for which $W(x, y) \geq \ell$ for all $x, y \in[0,1]$.

For any $\delta \geq 0$, we say that two graphons $W_0, W_1$ are a $\delta$-\emph{exceptional} pair if
\begin{align}
    \int_{0}^1 \left| \frac{d_{W_0}(\phi(x))}{D(W_0)} - \frac{d_{W_1}(x)}{D(W_1)}  \right| ~dx \leq \delta,
\end{align}
for some measure-preserving bijection $\phi:[0, 1]\to [0, 1]$.  If a pair of
graphons is not $\delta$-exceptional, then we say that they are $\delta$-separated.

We define the following class of activation functions:
\begin{definition}[Nice activation functions]
    We define $\Act$ to be the class of activation functions $\sigma:\R\to\R$ satisfying
    the following conditions:
    \begin{itemize}
        \item
            $\sigma \in C^2$.
        \item
            $\sigma(0) = 0$, $\sigma'(0) = 1$ and $\sigma'(x) \leq 1$ for all $x$.    
    \end{itemize}
\end{definition}
For simplicity, in Theorems~\ref{thm:convergence-result} and \ref{thm:prob-err-lower-bound} below, we will consider activations in the above class; however,
some of the conditions may be relaxed without inducing changes to our results:
in particular, we may remove the requirement that $\sigma'(0) = 1$, and
we may relax $\sigma'(x) \leq 1$ for all $x$ to only hold for $x$ in some
constant-length interval around $0$.
This expanded class includes activation functions such as $\sigma(x) = \tanh(x)$
 and the \emph{swish} and \emph{SELU} functions:
 \begin{itemize}
    \item
        swish~\cite{Hendrycks2017BridgingNA}: $\sigma(x) = \frac{x}{1+e^{-x}}$ 
    \item
        SELU~\cite{klambauer2017self}: 
            $\sigma(x) =  I[x \leq 0] (e^{x}-1) + I[x > 0] x$.
            
 \end{itemize}
 
We also make the following stipulation about the parameters of the GCN:
the initial embedding matrices $\hat{M}^{(b,0)}$ (with $b \in \{0, 1\}$) 
and weight matrices
$\{ W^{(j)} \}_{j=0}^{K}$ satisfy
\begin{align}
    \left\| M^{(b,0)T} \right\|_{op,\infty} 
    \cdot \prod_{j=0}^K \| W^{(j)T} \|_{op,\infty} \leq C,
\end{align}
and
    $\sum_{j=0}^K \| W^{(j)T} \|_{op,\infty} \leq E$,
for some fixed positive constants $C$ and $E$.

\AM{Revise this theorem statement!}
\begin{theorem}[Convergence of embedding vectors for a large class of
graphons and for a family of nonlinear activations]
    \label{thm:convergence-result}
    Let $W_0, W_1$ denote two $\delta$-exceptional graphons, for some fixed
    $\delta \geq 0$.
    
    
    Let $K$ satisfy $D\log n < K$, for some large enough constant $D > 0$
    that is a function of $W_0$ and $W_1$.  
    Consider the GCN with $K$
    layers and output embedding matrix $\hat{M}^{(K)}$, with the additional
    properties stated before the theorem.
    
    Suppose that $\delta > 0$.
    Then, in \emph{any} coupling of the graphs $G^{(0)} \sim W_0, G^{(1)} \sim W_1$, as $n \to \infty$, we have that the embedding vectors $\hat{H}^{(0,K)}$ and
    $\hat{H}^{(1,K)}$ satisfy
    \begin{align}
        \| \hat{H}^{(0,K)} - \hat{H}^{(1,K)} \|_{\infty} \leq  \frac{\delta}{n}(1 + O(1/\sqrt{n}))
        \label{expr:embeddings-convergence-bound-delta-nonzero}
    \end{align}
    with high probability.  
    
    If $\delta = 0$, then we have
    \begin{align}
        \| \hat{H}^{(0,K)} - \hat{H}^{(1,K)} \|_{\infty} \leq  O(n^{-3/2 + \text{const}}),
        \label{expr:embeddings-convergence-bound-delta-zero}
    \end{align}
    and for a $1-o(1)$-fraction of coordinates $i$, 
    $
        | \hat{H}^{(0,K)}_{i} - \hat{H}^{(1,K)}_{i} | = O(1/n^2).
    $
\end{theorem}
\begin{remark}
    The convergence bounds (\ref{expr:embeddings-convergence-bound-delta-nonzero})
    and (\ref{expr:embeddings-convergence-bound-delta-zero}) should be interpreted
    in light of the fact that the embedding vectors have entries on the order
    of $\Theta(1/n)$.
\end{remark}

Theorem~\ref{thm:convergence-result} can be translated, with some effort, to
the following result.

\begin{theorem}[Probability of error lower bound]
    \label{thm:prob-err-lower-bound}
    \AM{Make sure that the labels are right!}
    Consider again the setting of Theorem~\ref{thm:convergence-result}.
    Furthermore, 
    suppose that $\epsres > \frac{\delta}{2n}$.  Let $K$ additionally satisfy
    $K \ll n^{1/2 - \epsilon_0}$, for an arbitrarily small fixed $\epsilon_0 > 0$.
    Then there exist two sequences
    $\{ \Gr_{0,n} \}_{n=1}^{\infty}, \{ \Gr_{1,n} \}_{n=1}^{\infty}$
    of random graph models such that
    \begin{itemize}
        \item
            with probability $1$, samples $G_{b,n} \sim \Gr_{b,n}$ converge in cut
            distance to $W_b$,
        \item
            When $\delta > 0$,
            the probability of error of any test in distinguishing between $W_0$
            and $W_1$ based on $H^{(b,K)}$, the $\epsres$-uniform perturbation of
            $\hat{H}^{(b, K)}$, is at least
            \begin{align}
                \left( 1 - \frac{\delta}{2\epsres n} \right)^{n}
                \label{expr:err-prob-delta-nonzero}
            \end{align}
            \AM{Explain what this means in more detail.}
            
            When $\delta = 0$, the probability of error lower bound becomes
            \begin{align}
                \exp\left( - \frac{\text{const}}{\epsres \cdot n} \right).
                \label{expr:err-prob-delta-zero}
            \end{align}
            
        
    \end{itemize}   
\end{theorem}

\begin{remark}
    When $\epsres = \Theta(1/n)$ and $\delta = \Omega(1)$, the error probability lower bound (\ref{expr:err-prob-delta-nonzero}) is exponentially decaying to $0$.  On
    the other hand, when $\epsres \gg 1/n$ and $\delta = \Omega(1)$, it becomes
    $\exp\left( -\frac{\delta}{2\epsres}\right)(1+o(1))$,
    which is $\Theta(1)$.  
    
    When $\delta = 0$ and $\epsres = \Omega(1/n)$, the probability of error
    lower bound in (\ref{expr:err-prob-delta-zero}) is $\Omega(1)$.
\end{remark}
\AM{End theorem statement revision!}

We next turn to a positive result demonstrating the distinguishing capabilities
of very simple, linear GCNs.
\begin{theorem}[Distinguishability result]
    Let $W_0, W_1$ denote two $\delta$-separated graphons. 
    Then there exists a test that distinguishes with probability $1 - o(1)$ 
    between samples $G\sim W_0$
    and $G\sim W_1$ based on the $\epsres$-perturbed embedding vector from a
    GCN with $K$ layers, identity initial and weight matrices, and ReLU activation functions, provided that
    $K > D\log n$ for a sufficiently large $D$ and that $\epsres \leq \frac{\delta}{2n}$.
    \label{thm:achievability}
\end{theorem}

Finally, we exhibit a family of stochastic block 
models that are difficult to distinguish and are such that infinitely
many pairs of them have large cut distance.

To define the family of models, we consider the following density parameter
set: we pick a base point $P_* = (p_{*,1}, p_{*,2}, q_*)$ with all
positive numbers and then define
\begin{align*}
    &\Param  \\
    &= \left\{ P ~:~ (0, 0, 0) \prec P = P_* + \tau\cdot (\frac{1}{k_1}, \frac{k_1}{k_2^2}, \frac{-1}{k_2}) \preceq (1, 1, 1) \right\},
\end{align*}
where $\preceq$ is the lexicographic partial order, and $\tau \in \R$.
We have defined this parameter family because the corresponding SBMs
all have equal expected degree sequences.  

It may be checked that $\delta$ in Theorems~\ref{thm:convergence-result} and \ref{thm:prob-err-lower-bound}
is $0$ for pairs of graphons from $\Param$.  This gives the following result.
\begin{theorem}
    For any pair $W_0, W_1$ from the family of stochastic block models
    parameterized by $\Param$, there exists a $K > D\log n$, for some
    large enough positive constant $D$, such that the following statements
    hold:
    
    \paragraph{Convergence of embedding vectors}
    In any coupling of the graphs $G^{(0)} \sim W_0$ and $G^{(1)} \sim W_1$, as $n\to\infty$, we have that the embedding vectors
    $\hat{H}^{(0,K)}$ and $\hat{H}^{(1,K)}$ satisfy
    \begin{align}
        \| \hat{H}^{(0,K)} - \hat{H}^{(1,K)} \|_{\infty} = O(n^{-3/2 + \text{const}})
    \end{align}
    with probability $1 - e^{-\Theta(n)}$.
    \AM{Be more precise here!  Most of the coordinates differ by $O(n^{-2})$.}
    
    \paragraph{Probability of error lower bound}
    Let $K$ additionally satisfy $K \ll n^{1/2 - \epsilon_0}$, for an arbitrary small fixed
    $\epsilon_0 > 0$.  Then there exist two sequences
    $\{ \Gr_{0,n} \}_{n=1}^{\infty}$, 
    $\{ \Gr_{1,n} \}_{n=1}^{\infty}$ of random graph models
    such that
    \begin{itemize}
        \item
            with probability $1$, samples $G_{b,n} \sim \Gr_{b,n}$ converge in cut distance to $W_b$,
        \item
            the probability of error of any test in distinguishing between
            $W_0$ and $W_1$ based on $H^{(b,K)}$, the $\epsres$-uniform
            perturbation of $\hat{H}^{(b,K)}$, is lower
            bounded by
            $
                \exp\left( -\frac{C}{\epsres n} \right).
            $
    \end{itemize}
\end{theorem}

\section{Conclusions and future work}
We have shown conditions under which GCNs are information-theoretically
capable/incapable of distinguishing between sufficiently well-separated graphons.

It is worthwhile to discuss what lies ahead for the theory of
graph representation learning in relation to the problem of distinguishing
distributions on graphs.  As the present paper is a first step, we have left
several directions for future exploration.  
Most immediately, although we have proven impossibility results for GCNs with
nonlinear activation functions, we lack a complete
understanding of the benefits of more general
ways of incorporating nonlinearity.  We have shown
that architectures with too many layers cannot
be used to distinguish between graphons coming from a certain exceptional
class.  It would be of interest to determine if more general ways of incorporating
nonlinearity are 
able to generically distinguish between any sufficiently well-separated pair of 
graphons, whether or not they come from the exceptional class.  To this end, we are exploring
results indicating that replacing the random walk matrix $\hat{A}$ in the GCN
architecture with the transition matrix of a related Markov chain with the same
graph structure as the input graph $G$ results in a linear GCN that
is capable of distinguishing graphons generically.

Furthermore, a clear understanding of the role played by the embedding dimension
would be of interest.  In particular, we suspect that decreasing the embedding
dimension results in worse graphon discrimination performance.  
Moreover, a more precise
understanding of how performance parameters scale with the embedding dimension
would be valuable in GCN design.
Finally, we note that in many application domains, graphs are typically sparse.
Thus, we intend to generalize our theory to the sparse graph setting by replacing
graphons, which inherently generate dense graphs, with suitable nonparametric
sparse graph models, e.g., \emph{graphexes}.

\section{Acknowledgments}
This research was partially supported by grants from ARO W911NF-19-1026, ARO W911NF-15-1-0479, and ARO W911NF-14-1-0359 and the  Blue Sky Initiative from the College of Engineering at the University of Michigan.

\bibliographystyle{IEEEtran}
\bibliography{gnn-bib}

\end{document}